\documentclass[10pt,twocolumn,letterpaper]{article}
\usepackage[pagebackref=true,breaklinks=true,letterpaper=true,colorlinks,bookmarks=false]{hyperref}
\usepackage{cvpr}
\usepackage{times}
\usepackage{epsfig}
\usepackage{graphicx}
\usepackage{amsmath}
\usepackage{amssymb}
\usepackage{bbm}
\usepackage{comment}
\usepackage[ruled,lined]{algorithm2e}
\usepackage{makecell}
\usepackage{multirow}
\usepackage{subfigure}
\usepackage{enumitem}

\newcommand{\fancyname}{Next}

\cvprfinalcopy 

\def\cvprPaperID{2601} 

\ifcvprfinal\fi
\begin{document}

\title{Peeking into the Future: \\
Predicting Future Person Activities and Locations in Videos \\
}

\author{
Junwei Liang\textsuperscript{1}\thanks{Work partially done during a part-time research program at Google.} \qquad
Lu Jiang\textsuperscript{2} \qquad
Juan Carlos Niebles\textsuperscript{3,2} \qquad
Alexander Hauptmann\textsuperscript{1} \qquad 
Li Fei-Fei\textsuperscript{3,2}\vspace{.3em}\\
\textsuperscript{1}Carnegie Mellon University \qquad\qquad \textsuperscript{2}Google AI \qquad\qquad \textsuperscript{3}Stanford University\\
{\tt\small \{junweil,alex\}@cs.cmu.edu, lujiang@google.com, \{feifeili,jniebles\}@cs.stanford.edu} 
\vspace{-.5em}\\
}

\maketitle


\begin{abstract}
Deciphering human behaviors to predict their future paths/trajectories and what they would do from videos is important in many applications.
Motivated by this idea, this paper studies predicting a pedestrian's future path jointly with future activities. We propose an end-to-end, multi-task learning system utilizing rich visual features about human behavioral information and interaction with their surroundings. To facilitate the training, the network is learned with an auxiliary task of predicting future location in which the activity will happen.
Experimental results demonstrate our state-of-the-art performance over two public benchmarks on future trajectory prediction. Moreover, our method is able to produce meaningful future activity prediction in addition to the path. The result provides the first empirical evidence that joint modeling of paths and activities benefits future path prediction.~\footnote{Code and models are released at \url{https://next.cs.cmu.edu}}
\end{abstract}
\vspace{-6mm}


\setlength{\abovedisplayskip}{2pt} \setlength{\belowdisplayskip}{3pt}

\section{Introduction}


With the advancement in deep learning, systems now are able to analyze an unprecedented amount of rich visual information from videos to enable applications such as accident avoidance and smart personal assistance.
An important analysis is forecasting the future path of pedestrians, called future person path/trajectory prediction.
This problem has received increasing attention in the computer vision community~\cite{kitani2012activity,alahi2016social,gupta2018social}. It is regarded as an essential building block in video understanding because looking at the visual information from the past to predict the future is useful in many applications like self-driving cars, socially-aware robots~\cite{luber2010people}, \etc.

\begin{figure}[ht]
	\centering
		\includegraphics[width=0.47\textwidth]{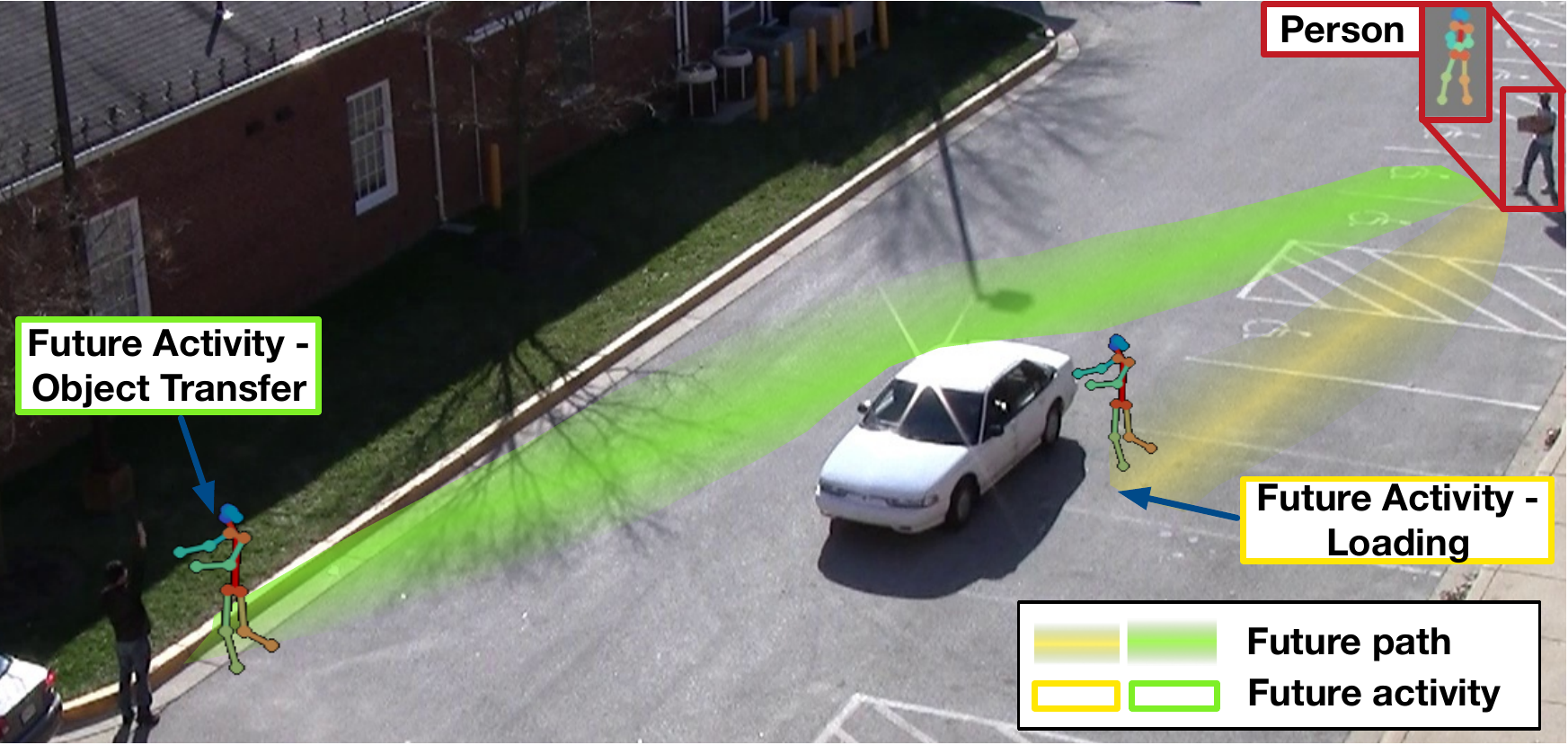} 
	\caption{Our goal is to jointly predict a person's future path and activity. The green and yellow line show two possible future trajectories and two possible activities are shown in the green and yellow boxes. Depending on the future activity, the person (top right) may take different paths, \eg the yellow path for ``loading'' and the green path for ``object transfer''. }
	\label{f1}
	\vspace{-5mm}
\end{figure}

Humans navigate through public spaces often with specific purposes in mind, ranging from simple ones like entering a room to more complicated ones like putting things into a car. Such intention, however, is mostly neglected in existing work. 
Consider the example in Fig.~\ref{f1}, the person (at the top-right corner) might take different paths depending on their intention, \eg, they might take the green path to~\emph{transfer object} or the yellow path to~\emph{load object into the car}. Inspired by this, this paper is interested in modeling the future path jointly with such intention in videos. 
We model the intention in terms of a predefined set of 29 activities provided by NIST such as ``loading'', ``object transfer'', \etc. See Table~\ref{tab:class} for the full list.

The joint prediction model can have two benefits. First, learning the activity together with the path may benefit the future path prediction. 
Intuitively, humans are able to read from others' body language to anticipate whether they are going to cross the street or continue walking along the sidewalk. After understanding these behaviors, humans can make better predictions. 
In the example of Fig.~\ref{f1}, the person is carrying a box, and the man at the bottom left corner is waving at the person. Based on common sense, we may agree that the person will take the green path instead of the yellow path. 
Second, the joint model advances the capability of understanding not only the future path but also the future activity by taking into account the rich semantic context in videos. 
This increases the capabilities of automated video analytics for social good such as real-time accident alerting, self-driving cars, and smart robot assistance. It may also have safety applications such as anticipating pedestrian movement at traffic intersections or a road robot helping humans transport goods to the trunk of a car. 
Note that our techniques focus on predicting a few seconds into the future, and should not be useful for non-routine activities.

To this end, we propose a multi-task learning model called~\emph{\fancyname} which has prediction modules for learning future paths and future activities simultaneously. 
As predicting future activity is challenging, we introduce two new techniques to address the issue. 
First, unlike most of the existing work~\cite{kitani2012activity, alahi2016social,gupta2018social,sadeghian2018sophie,manh2018scene,xie2018learning} which oversimplifies a person as a point in space, we encode a person through rich semantic features about visual appearance, body movement and interaction with the surroundings,  motivated by the fact that humans derive such predictions by relying on similar visual cues.
Second, to facilitate the training, we introduce an auxiliary task for future activity prediction, \ie activity location prediction.
In the auxiliary task, we design a discretized grid which we call the Manhattan Grid as location prediction target for the system.  
Experiments show that the auxiliary task improves the accuracy of future path prediction.

To the best of our knowledge, our work is the first on joint future path and activity prediction in streaming videos, and more importantly the first to demonstrate such joint modeling can considerably improve the future path prediction. 
We empirically validate our model on two benchmarks: ETH \& UCY~\cite{pellegrini2010improving,lerner2007crowds}, and ActEV/VIRAT~\cite{oh2011large,2018trecvidawad}. 
Experimental results show that our method outperforms state-of-the-art baselines, achieving the best-published result on two common benchmarks and producing additional prediction about the future activity.
To summarize, the contributions of this paper are threefold:
\textbf{(i)} We conduct a pilot study on joint future path and activity prediction in videos. We are the first to empirically demonstrate the benefit of such joint learning.
\textbf{(ii)} We propose a multi-task learning framework with new techniques to tackle the challenge of joint future path and activity prediction.
\textbf{(iii)} Our model achieves the best-published performance on two public benchmarks. Ablation studies are conducted to verify the contribution of the proposed sub-modules.

\begin{figure*}[!t]
	\centering
		\includegraphics[width=1.0\textwidth]{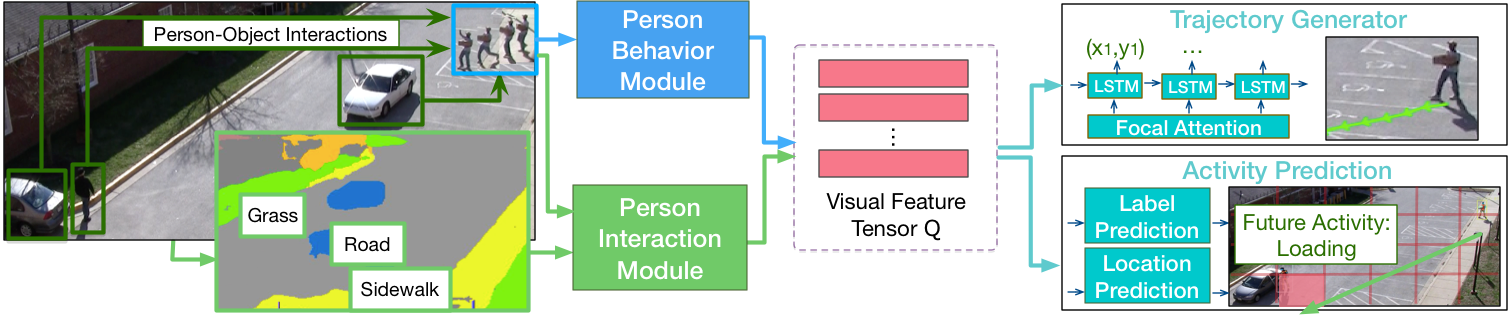}
	\caption{Overview of our model. Given a sequence of frames containing the person for prediction, our model utilizes person behavior module and person interaction module to encode rich visual semantics into a feature tensor. We propose novel person interaction module that takes into account both person-scene and person-object relations for joint activities and locations prediction.}  
	\label{f2}
	\vspace{-5mm}
\end{figure*}


\section{Related Work}

\noindent\textbf{Person-person models for trajectory prediction.} 
Person trajectory prediction models try to predict the future path of people, mostly pedestrians. 
A large body of work learns to predict person path by considering human social interactions and behaviors in crowded scene~\cite{xu2018encoding,yi2016pedestrian}.
Zou \etal in~\cite{zou2018understanding} learned human behaviors in crowds by imitating a decision-making process. 
Social-LSTM~\cite{alahi2016social} added social pooling to model nearby pedestrian trajectory patterns. 
Social-GAN~\cite{gupta2018social} added adversarial training on Social-LSTM to improve performance. 
Different from these previous work, we represent a person by rich visual features instead of simply considering a person as points in the scene. 
Meanwhile we use \emph{geometric relation} to explicitly model the person-scene interaction and the person-object relations, which have not been used in previous work.

\noindent\textbf{Person-scene models for trajectory prediction.} 
A number of works focused on learning the effects of the physical scene, \eg, people tend to walk on the sidewalk instead of grass. 
Kitani \etal in ~\cite{kitani2012activity} used Inverse Reinforcement Learning to forecast human trajectory. 
Xie \etal in ~\cite{xie2018learning} considered pedestrian as ``particles'' whose motion dynamics are modeled within the framework of Lagrangian Mechanics. Scene-LSTM~\cite{manh2018scene} divided the static scene into Manhattan Grid and predict pedestrian's location using LSTM. 
CAR-Net~\cite{jaipuria2018transferable} proposed an attention network on top of scene semantic CNN to predict person trajectory.
SoPhie~\cite{sadeghian2018sophie} combined deep neural network features from scene semantic segmentation model and generative adversarial network (GAN) using attention to model person trajectory.
A disparity to ~\cite{sadeghian2018sophie} is that we explicitly pool scene semantic features around each person at each time instant so that the model can directly learn from such interactions.

\noindent\textbf{Person visual features for trajectory prediction.} 
Some recent works have attempted to predict person path by utilizing individual's visual features instead of considering them as points in the scene. Kooij \etal in~\cite{kooij2014context} looked at pedestrian's faces to model their awareness to predict whether they will cross the road using a Dynamic Bayesian Network in dash-cam videos. Yagi \etal in ~\cite{yagi2018future} used person keypoint features with a convolutional neural network to predict future path in first-person videos. 
Different from these works, we consider rich visual semantics for future prediction that includes both the person behavior and their interactions with soundings .

\noindent\textbf{Activity prediction/early recognition.}
Many works have been proposed to anticipate future human actions using Recurrent Neural Network (RNN). \cite{ma2016learning} and \cite{aliakbarian2017encouraging} proposed different losses to encourage LSTM to recognize actions early in internet videos. Srivastava \etal in~\cite{srivastava2015unsupervised} utilized unsupervised learning with LSTM to reconstruct and predict video representations.
Another line of works is anticipating human activities in robotic vision ~\cite{koppula2016anticipating,jain2015car}.
Our work differs in that both person behavior and person interaction modeling are used for joint activity and trajectory prediction.

\noindent\textbf{Multiple cues for tracking/group activity recognition.} There are previous works that take into account multiple cues in videos for tracking~\cite{jain2016structural, sadeghian2017tracking} and group activity recognition~\cite{choi2014understanding, shu2015joint, shu2017cern}. Our work differs in that rich visual features and focal attention are used for joint person path and activity prediction. Meanwhile, our work utilizes novel activity location prediction (see Section~\ref{sec:cls_rg}) to bridge the two tasks.


\section{Approach}

Humans navigate through spaces often with specific purposes in mind. Such purposes may considerably orient the future trajectory/path. This motivates us to study the future path prediction jointly with the intention. In this paper, we model the intention in terms of a predefined set of future activities such as ``walk'', ``open\_door'', ``talk'', \etc.

\noindent\textbf{Problem Formulation}: Following~\cite{alahi2016social,gupta2018social,sadeghian2018sophie}, we assume each scene is first processed to obtain the spatial coordinates of all people at different time instants. Based on the coordinates, we can automatically extract their bounding boxes. Our system observes the bounding box of all the people from time 1 to $T_{obs}$, and objects if there are any, and predicts their positions (in terms of $xy$-coordinates) for time $T_{obs+1}$ to $T_{pred}$, meanwhile estimating the possibilities of future activity labels at time $T_{pred}$.

\subsection{Network Architecture} \label{network}
Fig.~\ref{f2} shows the overall network architecture of our \emph{\fancyname} model. 
Unlike most of the existing work~\cite{kitani2012activity, alahi2016social,gupta2018social,sadeghian2018sophie,manh2018scene,xie2018learning} which oversimplifies a person as a point in space, our model employs two modules to encode rich visual information about each person's behavior and interaction with the surroundings. In summary, it has the following key components:

\noindent\textbf{Person behavior module} extracts visual information from the behavioral sequence of the person. 


\noindent\textbf{Person interaction module} looks at the interaction between a person and their surroundings. 

\noindent\textbf{Trajectory generator} summarizes the encoded visual features and predicts the future trajectory by the LSTM decoder with focal attention~\cite{liang2018focal}. 


\noindent\textbf{Activity prediction} utilizes rich visual semantics to predict the future activity label for the person.
In addition, we divide the scene into a discretized grid of multiple scales, which we call the Manhattan Grid, to compute classification and regression for robust activity location prediction. 

In the rest of this section, we will introduce the above modules and the learning objective in details.


\subsection{Person Behavior Module} \label{sec:person-behavior}
\begin{figure}[t]
	\centering
		\includegraphics[width=0.47\textwidth]{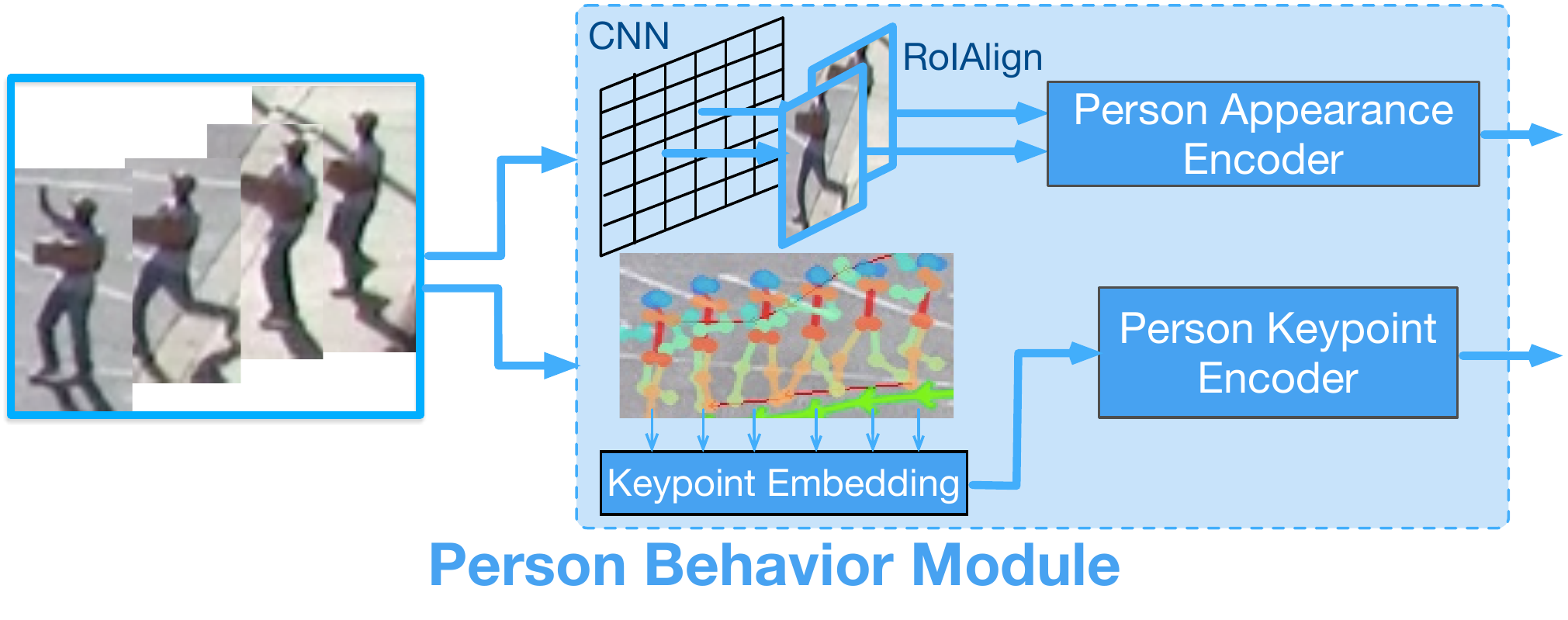}
			\vspace{-2mm}
	\caption{We show the person behavior module given a sequence of person frames. We extract person appearance features and pose features to model the changes of a person's behavior. See Section~\ref{sec:person-behavior}.}
	\label{behavior}
	\vspace{-5mm}
\end{figure}

This module encodes the visual information about every individual in a scene. As opposed to oversimplifying a person as a point in space, we model the person's the appearance and body movement. To model appearance changes of a person, we utilize a pre-trained object detection model with ``RoIAlign''~\cite{he2017mask} to extract fixed size CNN features for each person bounding box. See Fig.~\ref{behavior}. For every person in the scene, we average the feature along the spatial dimensions and feed them into an LSTM encoder. Finally, we obtain a feature representation of ${T_{obs} \times d}$, where $d$ is the hidden size of the LSTM. 

To capture the body movement, we utilize a person keypoint detection model trained on MSCOCO dataset~\cite{fang2017rmpe} to extract person keypoint information. We apply the linear transformation to embed the keypoint coordinates before feeding into the LSTM encoder. The shape of the encoded feature has the shape of ${T_{obs} \times d}$. These appearance and movement features are commonly used in a wide variety of studies and thus do not introduce new concern on machine learning fairness.




\subsection{Person Interaction Module} \label{sec:person-scene}
This module looks at the interaction between a person and their surroundings, \ie person-scene and person-objects interactions.

\noindent\textbf{Person-scene.} 
To encode the nearby scene of a person, we first use a pre-trained scene segmentation model~\cite{deeplabv3plus2018} to extract pixel-level scene semantic classes for each frame. We use totally $N_s=10$ common scene classes, such as roads, sidewalks, \etc. 
The scene semantic features are integers (class indexes) of the size $T_{obs} \times h \times w$, where $h,w$ are the spatial resolution. We first transform the integer tensor into $N_s$ binary masks (one mask for each class), and average along the temporal dimension. This results in $N_s$ real-valued masks, each of the size of $h \times w$. We apply two convolutional layers on the mask feature with a stride of 2 to get the \emph{scene CNN features} in two scales.

\begin{figure}[t]
	\centering
		\includegraphics[width=0.47\textwidth]{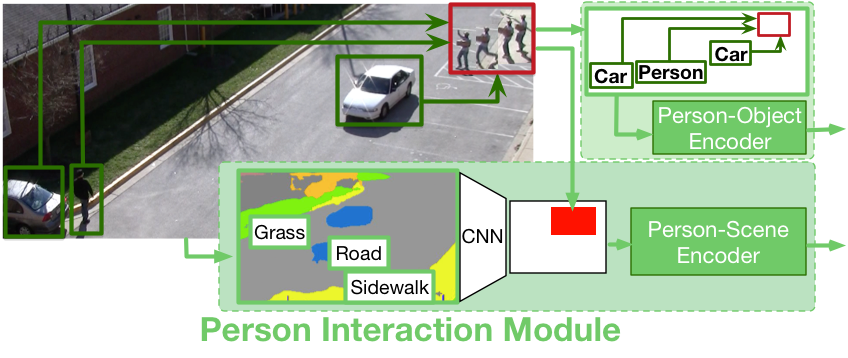}
		\vspace{-2mm}
	\caption{We show the person interaction module which includes person-scene and person-objects modeling. For person-objects modeling, given the person sequence as the red box in the video frame, we extract the spatial relations between the person and other objects at each time instant. For person-scene modeling, surrounding scene semantic features are pooled around the person into the encoder. See Section~\ref{sec:person-scene}.}
	\label{person-scene}
	\vspace{-3mm}
\end{figure}

Given a person's $xy$-coordinate, we pool the scene features at the person's current location from the convolution feature map. As the example shown at the bottom of Fig.~\ref{person-scene}, the red part of the convolution feature is the discretized location of the person at the current time instant.
The receptive field of the feature at each time instant, \ie the size of the spatial window around the person which the model looks at, depends on which scale is being pooled from and the convolution kernel size. 
In our experiments, we set the scale to 1 and the kernel size to 3, which means our model looks at the 3-by-3 
surrounding area of the person at each time instant. 
The person-scene representation for a person is in $\mathbb{R}^{T_{obs} \times C}$, where $C$ is the number of channels in the convolution layer. We feed this into a LSTM encoder in order to capture the temporal information and get the final person-scene features in $\mathbb{R}^{T_{obs} \times d}$.

\noindent\textbf{Person-objects.} Unlike previous work~\cite{alahi2016social,gupta2018social} which relies on LSTM hidden states to model nearby people, our module explicitly models the \textit{geometric relation} and the \textit{object type} of all the objects/persons in the scene. At any time instant, given the observed box of a person  $(x_{b}, y_{b}, w_{b}, h_{b})$ and $K$ other objects/persons in the scene $\{(x_{k}, y_{k}, w_{k}, h_{k}) | k \in [1, K] \}$, we encode the geometric relation into $\mathcal{G} \in \mathbb{R}^{K \times 4}$, the $k$-th row of which equals to:
\begin{equation}
\small
\mathcal{G}_{k} =  [\log(\frac{|x_{b} - x_{k}|}{w_{b}}), \log(\frac{|y_{b} - y_{k}|}{h_{b}}), \log(\frac{w_{k}}{w_{b}}), \log(\frac{h_{k}}{h_{b}})]
\end{equation}
This encoding computes the geometric relation in terms of the geometric distance and the fraction box size. We use a logarithmic function to reflect our observation that human trajectories are more likely to be affected by close-by objects or people. This encoding has been proven effective in object detection~\cite{hu2018relation}.

For the object type, we simply use one-hot encoding to get the feature in $\mathbb{R}^{K \times N_o}$, where $N_o$ is the total number of object classes. We then embed the geometric features and the object type features at the current time into $d_e$-dimensional vectors and feed the embedded features into an LSTM encoder to obtain the final feature of the shape $T_{obs} \times d$.


As shown in the example from Fig.~\ref{person-scene}, the person-objects feature can capture how far away the person is to the other person and the cars (with respect to their own height). The person-scene feature can capture whether the person is near the sidewalk or grass.
We feed this information to the model with the hope of learning things like a person walks more often on the sidewalk than the grass and tends to avoid bumping into cars.


\subsection{Trajectory Generation with Focal Attention}\label{sec:focal_attention}
As discussed, the above four types of visual features, \ie appearance, body movement, person-scene, and person-objects, are encoded by separate LSTM encoders into the same dimension. Besides, given a person's trajectory output from the last time instant, we extract the trajectory embedding by
\begin{equation}
    e_{t-1} = \tanh\{W_{e} [x_{t-1}, y_{t-1}]\} + b_e \in \mathbb{R}^{d},
\end{equation}
where $[x_{t-1}, y_{t-1}]$ is the trajectory prediction of time $t-1$ and $W_e, b_e$ are learnable parameters. We then feed the embedding $e_{t-1}$ into another LSTM encoder for the trajectory. The hidden states of all encoders are packed into a tensor named $Q \in \mathbb{R}^{M \times T_{obs} \times d}$, where $M=5$ denotes the total number of features and $d$ is the hidden size of the LSTM.

Following~\cite{gupta2018social}, we use an LSTM decoder to directly predict the future trajectory in the $xy$-coordinate. The hidden state of this decoder is initialized using the last state of the person's trajectory LSTM encoder. At each time instant, the $xy$-coordinate will be computed from the decoder state $h_t = \text{LSTM}(h_{t-1}, [e_{t-1}, \tilde{q}_t])$ and by a fully connected layer. $\tilde{q}_t$ is an important attended feature vector which summarizes salient cues in the input features $Q$. We employ an effective focal attention~\cite{liang2018focal} to this end. It was originally proposed to carry out multimodal inference over a sequence of images for visual question answering. The key idea is to project multiple features into a space of correlation, where discriminative features can be easier to capture by the attention mechanism. 

To do so, we compute a correlation matrix $S^{t} \in \mathbb{R}^{M \times T_{obs}}$ at every time instant $t$, where each entry $S^t_{ij} = h_{t-1}^{\top} \cdot Q_{ij:}$ is measured using the dot product similarity and $:$ is a slicing operator that extracts all elements from that dimension. Then we compute two focal attention matrices: 
\begin{align}
\small
    A^t &= \text{softmax}(\max_{i=1}^{M} S^t_{i:}) \in \mathbb{R}^{M} \\
    B^t &= [\text{softmax}(S^t_{1:}), \cdots, \text{softmax}(S^t_{M:})] \in \mathbb{R}^{M \times T_{obs}} 
\end{align}
Then the attended feature vector is given by:
\begin{equation} \label{eq:focal}
\tilde{q}_t = \sum_{j=1}^M A^t_{j} \sum_{k=1}^{T_{obs}} B^t_{jk} Q_{jk:} \in \mathbb{R}^{d}
\end{equation}
As shown, the focal attention models the correlation among different features and summarizes them into a low-dimensional attended vector. Section~\ref{sec:exp} show its benefit in our experiments.



\subsection{Activity Prediction}\label{sec:cls_rg}
Since the trajectory generation module outputs one location at a time, errors may accumulate across time and the final destination would deviate from the actual location. 
Using the wrong location for activity prediction may lead to bad accuracy.
To counter this disadvantage, we introduce an auxiliary task, \ie activity location prediction, in addition to predicting the future activity label of the person. We describe the two prediction modules in the following.

\begin{figure}[t]
	\centering
		\includegraphics[width=0.47\textwidth]{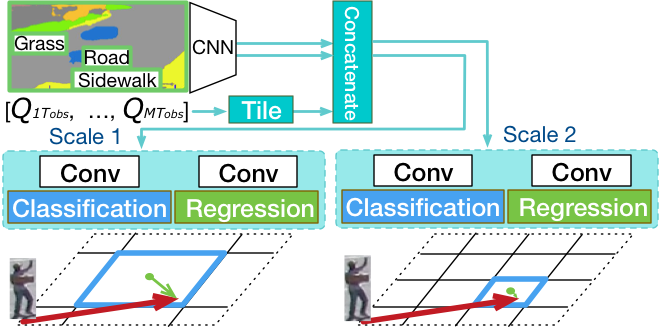}
	\caption{Activity location prediction with classification and regression on the multi-scale Manhattan Grid. See Section~\ref{sec:cls_rg}.}
	\label{grid_loss}
	\vspace{-5mm}
\end{figure}

\noindent\textbf{Activity location prediction with the Manhattan Grid.}
To bridge the gap between trajectory generation and activity label prediction, we propose an activity location prediction module to predict the final location of where the person will engage in the future activity. 
The activity location prediction includes two tasks, \emph{location classification} and \emph{location regression}.
As illustrated in Fig.~\ref{grid_loss}, we first divide a video frame into a discretized $h \times w$ grid, namely \emph{Manhattan Grid}, and learn to classify the correct grid block and at the same time to regress from the center of that grid block to the actual location. 
Specifically, the aim for the classification task is to predict the correct grid block in which the final location coordinates reside. 
After classifying the grid block, the aim for the regression task is to predict the deviation of the grid block center (Green Dot in the figure) to the final location coordinate (the end of Green Arrow).
The reason for adding the regression task are: (i) it will provide more precise locations than just a grid block area; (ii) it is complementary to the trajectory prediction which requires $xy$-coordinates localization.
We repeat this process on the Manhattan Grid of different scales and use separate prediction heads to model them. These prediction heads are trained end-to-end with the rest of the model. 
Our idea is partially inspired by the region proposal network~\cite{ren2015faster} and our intuition is that similar to object detection problem, we need accurate localization using multi-scale features in a cost-efficient way.

As shown in Fig.~\ref{grid_loss}, we first concatenate the scene CNN features (see Section~\ref{sec:person-scene}) with the last hidden state of the encoders (see Section~\ref{sec:focal_attention}). For compatibility, we tile the hidden state $Q_{:T_{obs}:}$ along the height and width dimension resulting in a tensor of the size $M \times d \times w \cdot h$, where $w \cdot h$ is the total number of the grid blocks. The hidden state contains rich information from all encoders and allow gradients flow smoothly through from prediction to feature encoders.

The concatenated features are fed into two separate convolution layers for classification and regression. The convolution output for grid classification $\mathsf{cls}_{grid} \in \mathbb{R}^{w \cdot h \times 1}$ indicates the probability of each grid block being the correct destination. In comparison, the convolution output for grid regression $\mathsf{rg}_{grid} \in \mathbb{R}^{w \cdot h \times 2}$ denotes the deviation, in the $xy$-coordinates, between the final destination and every grid block center. A row of $\mathsf{rg}_{grid}$ represents the difference to a grid block, calculated from $[x_t-x_{ci}, y_t-y_{ci}]$ where $(x_t, y_t)$ denotes the predicted location and $(x_{ci},y_{ci} )$ is the center of the $i$-th grid block. The ground truth for the grid regression can be computed in a similar way. During training, only the correct grid block receives gradients for regression.
Recent work~\cite{manh2018scene} also incorporates the grid for location prediction. Our model differs in that we link grid locations to scene semantics, and use a classification layer and a regression layer together to make more robust predictions.

\noindent\textbf{Activity label prediction.}
Given the encoded visual observation sequence, the activity label prediction module predicts the future activity at time instant $T_{pred}$. We compute the future $N_{a}$ activity probabilities using the concatenated last hidden states of the encoders:
\begin{equation}
    \mathsf{cls}_{act} = \text{softmax}(W_{a} \cdot [Q_{1T_{obs}:}, \cdots, Q_{MT_{obs}:}])
\end{equation}
where $W_a$ is a learnable weight. The future activity of a person could be multi-class, \eg a person could be ``walking'' and ``carrying'' at the same time.

\vspace{-2mm}
\subsection{Training}
The entire network is trained end-to-end by minimizing a multi-task objective. The primary loss is the common $L_2$ loss between the predicted future trajectories and the ground-truth trajectories~\cite{manh2018scene,gupta2018social,sadeghian2018sophie}. The loss is summed into $L_{xy}$ over all persons from $T_{obs+1}$ to $T_{pred}$.

The second category of loss is the activity location classification and regression loss discussed in Section~\ref{sec:cls_rg}. We have $L_{grid\_cls} \!=\! \!\sum_{i=1}^N \! \text{ce}(\mathsf{cls}^i_{grid},\mathsf{cls}^{\ast i}_{grid})$, where $\mathsf{cls}^{\ast i}_{grid}$ is the ground-truth final location grid block ID for the $i^{th}$ training trajectory. 
Likewise $L_{grid\_reg} \!=\! \!\sum_{i=1}^N \text{smooth}_{L_1}(\mathsf{rg}^i_{grid},\mathsf{rg}^{\ast i}_{grid})$ and $\mathsf{rg}^{\ast i}_{grid}$ is the ground-truth difference to the correct grid block center. This loss is designed to bridge the gap between the trajectory generation task and activity label prediction task.

The third loss is for activity label prediction.
We employ the cross-entropy loss: $L_{act} = \sum_{i=1}^N \text{ce}(\mathsf{cls}^i_{act},\mathsf{cls}^{\ast i}_{act})$.
The final loss is then calculated from: 
\begin{equation}
    L = L_{xy} + \lambda(L_{grid\_cls} + L_{grid\_reg}) + L_{act}
\end{equation}
We use a balance controller $\lambda = 0.1$ for location destination prediction to offset their higher loss values during training.\label{sec:approach}


\vspace{-2mm}
\section{Experiments}\label{sec:exp}

We evaluate the proposed~\emph{\fancyname}~model on two common benchmarks for future path prediction: ETH~\cite{pellegrini2010improving} and UCY~\cite{lerner2007crowds}, and ActEV/VIRAT~\cite{2018trecvidawad,oh2011large}. 
We demonstrate that our model performs favorably against the state-of-the-art models on this challenging task. The source code and models will be made available to the public.


\subsection{ActEV/VIRAT}\label{sec:exp-virat}

\noindent\textbf{Dataset \& Setups.} ActEV/VIRAT~\cite{2018trecvidawad} is a public dataset released by NIST in 2018 for activity detection research in streaming video ({\footnotesize \url{https://actev.nist.gov/}}). This dataset is an improved version of VIRAT~\cite{oh2011large}, with more videos and annotations. It includes 455 videos at 30 fps from 12 scenes, more than 12 hours of recordings. Most of the videos have a high resolution of 1920x1080. 
We use the official training set for training and the official validation set for testing.

Following~\cite{alahi2016social, gupta2018social,sadeghian2018sophie}, the models observe 3.2 seconds (8 frames) of every person and predict the future 4.8 seconds (12 frames) of person trajectory, we downsample the videos to 2.5 fps and extract person trajectories using the code released in~\cite{gupta2018social}. Since we do not have the homographic matrix, we use the pixel values for the trajectory coordinates as it is done in~\cite{yagi2018future}. 

\noindent\textbf{Evaluation Metrics.} Following prior work~\cite{alahi2016social,gupta2018social,sadeghian2018sophie}, we use two error metrics for person trajectory prediction: 

\noindent i) \textit{Average Displacement Error} (ADE): The average Euclidean distance between the ground truth coordinates and the prediction coordinates over all time instants,
\begin{equation}
    \text{ADE} = \frac{ \sum^{N}_{i=1} \sum^{T_{pred}}_{t=1} \lVert \tilde{Y}^i_t - Y^i_t \rVert_{2}}{N*T_{pred}}
\end{equation}

\noindent ii) \textit{Final Displacement Error} (FDE): The euclidean distance between the predicted points and the ground truth point at the final prediction time instant $T_{pred}$,
\begin{equation}
    \text{FDE} = \frac{ \sum^{N}_{i=1} \lVert \tilde{Y}^i_{T_{pred}} - Y^i_{T_{pred}} \rVert_{2}}{N}
\end{equation}
The errors are measured in the pixel space on ActEV/VIRAT whereas in meters on ETH and UCY. 
For future activity prediction, we use mean average precision (mAP).

\noindent\textbf{Baseline methods.} We compare our method with the two simple baselines and two recent methods: \textbf{\textit{Linear}} is a single layer model that predicts the next coordinates using a linear regressor based on the previous input point. \textbf{\textit{LSTM}} is a simple LSTM encoder-decoder model with coordinates input only. \textbf{\textit{Social LSTM}}~\cite{alahi2016social}: We train the social LSTM model to directly predict trajectory coordinates instead of Gaussian parameters. \textbf{\textit{SGAN}}~\cite{gupta2018social}: We train two model variants (PV \& V) detailed in the paper using the released code from Social-GAN~\cite{gupta2018social} {\footnotesize (\url{https://github.com/agrimgupta92/sgan/})}.

Aside from using a single model at test time, Gupta \etal~\cite{gupta2018social} also used 20 model outputs per frame and selected the best prediction to count towards the final performance. Following the practice, we train 20 identical models using random initializations and report the same evaluation results, which are marked ``20 outputs'' in Table~\ref{exp1}.


\begin{table}[t]
\small
\begin{tabular}{l|l||c|c|c|c}
\hline
&Method & ADE          & FDE          & {\scriptsize move\_ADE}     & {\scriptsize move\_FDE}      \\ \hline \hline
\multirow{6}{*}{\rotatebox[origin=c]{90}{\footnotesize{Single Model}}}&Linear                     & 32.19        & 60.92        & 42.82        & 80.18         \\ 
&LSTM & 23.98        & 44.97        & 30.55        & 56.25        \\ 
&Social LSTM  & 23.10        & 44.27        & 28.59        & 53.75       \\ 
&SGAN-PV  & 30.51        & 60.90        & 37.65        & 73.01         \\ 
&SGAN-V   & 30.48        & 62.17        & 35.41        & 68.77         \\ 
&Ours     & \textbf{17.99} & \textbf{37.24} & \textbf{20.34} & \textbf{42.54} \\
&Ours-Noisy     & 34.32 &  57.04  & 40.33 &	66.73 \\
\hline
\multirow{3}{*}{\rotatebox[origin=c]{90}{\scriptsize{20 Outputs}}}&SGAN-PV-20 & 23.11        & 41.81        & 29.80        & 53.04         \\ 
&SGAN-V-20  & 21.16        & 38.05        & 26.97        & 47.57       \\ 
&Ours-20                            & \textbf{16.00}        & \textbf{32.99}        & \textbf{17.97}        & \textbf{37.28}             \\ \hline
\end{tabular}
\caption{Comparison to baseline methods on the ActEV/VIRAT validation set. Top uses the single model output. Bottom uses 20 outputs. Numbers denote errors thus lower are better.}
\label{exp1}
\vspace{-5mm}
\end{table}

\noindent\textbf{Implementation Details.}
We use LSTM cell for both the encoder and decoder. 
The embedding size $d_e$ is set to 128, and the hidden sizes $d$ of encoder and decoder are both 256. 
Ground truth bounding boxes of persons and objects are used during the observation period (from time 1 to $T_{obs}$).
For person keypoint features, we utilize the pre-trained pose estimator from~\cite{fang2017rmpe} to extract 17 joints for each ground truth person box. 
For person appearance feature, we utilize the pre-trained object detection model FPN~\cite{lin2017feature} to extract appearance features from person bounding boxes. 
The scene semantic segmentation features are resized to (64, 36) and the scene convolution layers are set to have a kernel size of 3, a stride of 2 and the channel dimension is 64. 
We resize all videos to 1920x1080 and utilize two grid scales, 32x18 and 16x9. 
The activation function is $tanh$ if not stated otherwise and we do not use any normalization. 
For training, we use Adadelta optimizer~\cite{zeiler2012adadelta} with an initial learning rate of 0.1 and the dropout value is 0.3. 
We use gradient clipping of 10 and weight decay of 0.0001. 
For Social LSTM, the neighbor is set to 256 pixels as in~\cite{yagi2018future}. 
All baselines use the same embedding size and hidden size as our model, therefore all encoder-decoder models have about the same numbers of parameters. Other hyper-parameters we use for the baselines follow the ones in ~\cite{gupta2018social}.

\begin{figure*}[!t]
	\subfigure{\label{fig:comparea}}
    \subfigure{\label{fig:compareb}}
    \subfigure{\label{fig:comparec}}
    \subfigure{\label{fig:compared}}
	\centering
		\includegraphics[width=1.0\textwidth]{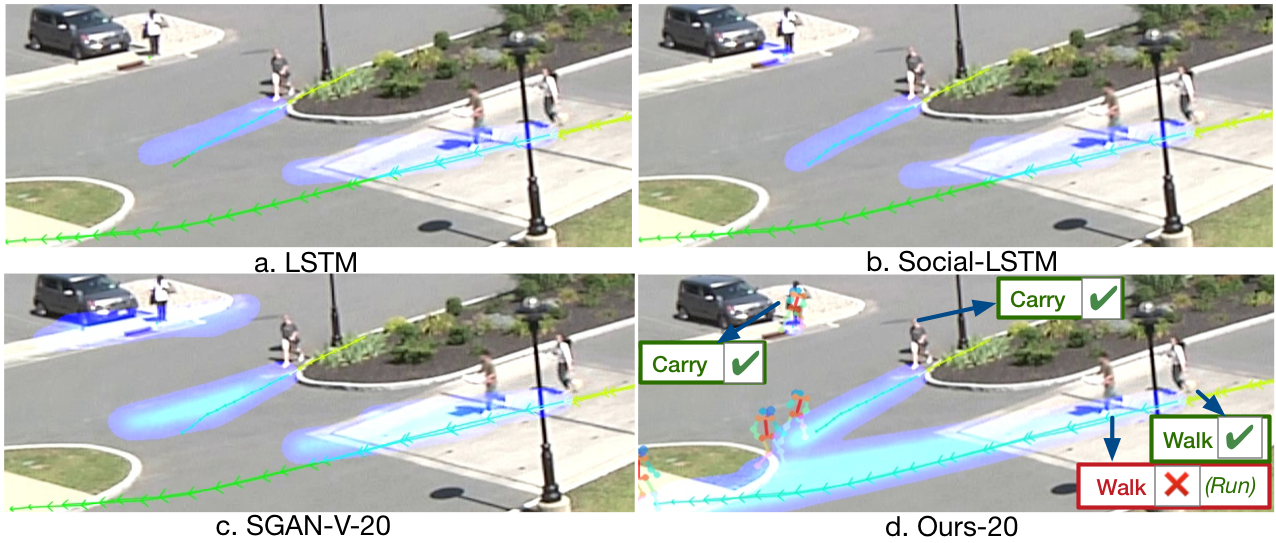} 
		\vspace{-5mm}
	\caption{(Better viewed in color.) Qualitative comparison between our method and the baselines. Yellow path is the observable trajectory and Green path is the ground truth trajectory during the prediction period. Predictions are shown as Blue heatmaps. Our model also predicts the future activity, which is shown in the text and with the person pose template.}  
	\label{fig:qualitative-compare}
		\vspace{-5mm}
\end{figure*}

\noindent\textbf{Main Results.} Table~\ref{exp1} lists the testing error, where the top part is the error of a single model output and the bottom shows the best result of 20 model outputs. The ``ADE'' and ``FDE'' columns summarize the error over all trajectories, and the last two columns further detail the subset trajectories of moving activities (``walk'', ``run'', and ``ride\_bike''). 
We report the mean performance of 20 runs of our single model at Row 7. The standard deviation on ``ADE'' metric is 0.043. Full numbers can be found in supplemental material.
As we see, our method performs favorably against other methods, especially in predicting the trajectories of moving activities. For example, our model outperforms Social-LSTM and Social-GAN by a large margin of 10 points in terms of the ``move\_FDE'' metric. The results demonstrate the efficacy of the proposed model and its state-of-the-art performance on future trajectory prediction.
Additionally, as a step towards real-world application, we train our model with noisy outputs from object detection and tracking during the observation period. For evaluation, following common practise in tracking~\cite{wu2015object}, for each trajectory, we assume the person bounding box location at time $1$ is close to the ground truth location, and we evaluate the model prediction using tracking inputs and other visual features from time $1$ to $T_{obs}$ as shown in Table~\ref{exp1} ``Ours-Noisy''.

\noindent\textbf{Qualitative analysis.} 
We visualize and compare our model outputs and the baselines in Fig.~\ref{fig:qualitative-compare}.
In each graph the yellow trajectories are the observable sequences of each person and the green trajectories are the ground truth future trajectories. The predicted trajectories are shown in the blue heatmap. To better visualize the predicted future activities of our method, we plot the person keypoint template for each predicted activity at the end of the predicted trajectory.
As we see, our method outputs more accurate trajectories for each person, especially for the two persons on the right that were about to accelerate their movement. Our method is also able to predict most of the activities correct except one (walk versus run).
Our model successfully predicts the activity ``carry'' and the static trajectory of the person near the car, while in Fig~\ref{fig:comparec}, SGAN predicts several moving trajectories in different directions. 

\begin{figure*}[!t]
	\subfigure{\label{fig:ourmodela}}
    \subfigure{\label{fig:ourmodelb}}
    \subfigure{\label{fig:ourmodelc}}
    \subfigure{\label{fig:ourmodeld}}
	\centering
		\includegraphics[width=1.0\textwidth]{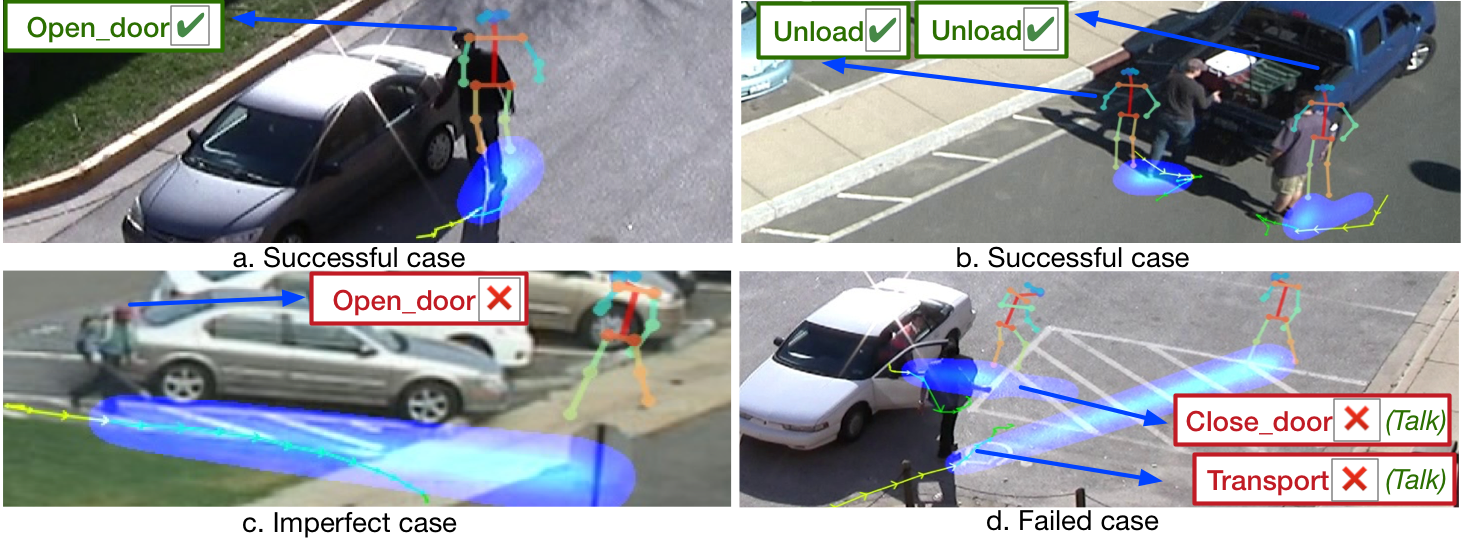}
		\vspace{-6mm}
	\caption{(Better viewed in color.) Qualitative analysis of our model. Please refer to Fig.~\ref{fig:qualitative-compare} for legends.}  
	\label{fig:qualitative}
	\vspace{-3mm}
\end{figure*}

We further provide a qualitative analysis of our model predictions. (i) Successful cases: In Fig~\ref{fig:ourmodela} and~\ref{fig:ourmodelb}, both the trajectory prediction and future activity prediction are correct. 
(ii) Imperfect case: In Fig~\ref{fig:ourmodelc}, although the trajectory prediction is mostly correct, our model predicts that the person is going to open the door of the car, given the observation that he is walking towards the side of the car. 
(iii) Failed case: In Fig~\ref{fig:ourmodeld}, our model fails to capture the subtle interactions between the two persons and predicts that they will go separate ways, while in fact they are going to stop and talk to each other. 

\begin{table}
\centering
\small
\begin{tabular}{l||c|c|c}
\hline
Method                             & ADE $\downarrow$  & FDE $\downarrow$  & Act mAP $\uparrow$\\ \hline
Our full model      & 17.91 & 37.11 & 0.192           \\ \hline
No p-behavior           & 18.99 & 39.82 & 0.139            \\ 
No p-interaction          & 18.83 & 39.35    & 0.163            \\ 
No focal attention                 & 19.93 & 42.08 & 0.144           \\ 
No act label loss                        & 19.48 & 41.45 & -                 \\
No act location loss              &19.07 & 39.91 &   0.152          \\
No multi-task               & 20.37 & 42.79     & -                        \\ \hline
\end{tabular}
\caption{Multi-task performance \& ablation experiments.}
\label{multi-task}
\vspace{-7mm}
\end{table}

\begin{table*}[]
\centering
\small
\begin{tabular}{l|l||c|c||c|c|c||c}
\hline
                   & Method& ETH         & HOTEL       & UNIV  *      & ZARA1       & ZARA2       & AVG         \\ \hline \hline
\multirow{4}{*}{\rotatebox[origin=c]{90}{\footnotesize{Single Model}}}&Linear & 1.33 / 2.94 & 0.39 / 0.72 & 0.82 / 1.59 & 0.62 / 1.21 & 0.77 / 1.48 & 0.79 / 1.59 \\ 
&LSTM                & 1.09 / 2.41 & 0.86 / 1.91 & \textbf{0.61} / \textbf{1.31} & \textbf{0.41} / \textbf{0.88} & 0.52 / 1.11 & 0.70 / 1.52 \\ 
&Alahi \etal~\cite{alahi2016social}         & 1.09 / 2.35 & 0.79 / 1.76 & 0.67 / 1.40 & 0.47 / 1.00 & 0.56 / 1.17 & 0.72 / 1.54 \\ 
&Ours-single-model   & \textbf{0.88} / \textbf{1.98} & \textbf{0.36} / \textbf{0.74} & 0.62 / 1.32 & 0.42 / 0.90 & \textbf{0.34} / \textbf{0.75} & \textbf{0.52} / \textbf{1.14} \\ \hline

\multirow{4}{*}{\rotatebox[origin=c]{90}{20 Outputs}}&Gupta \etal~\cite{gupta2018social}(V)         & 0.81 / 1.52 & 0.72 / 1.61 & 0.60 / 1.26 & 0.34 / 0.69 & 0.42 / 0.84 & 0.58 / 1.18 \\ 
&Gupta \etal~\cite{gupta2018social}(PV)        & 0.87 / 1.62 & 0.67 / 1.37 & 0.76 / 1.52 & 0.35 / 0.68 & 0.42 / 0.84 & 0.61 / 1.21 \\ 
&Sadeghian \etal~\cite{sadeghian2018sophie}             & \textbf{0.70} / \textbf{1.43} & 0.76 / 1.67 & \textbf{0.54} / \textbf{1.24} & \textbf{0.30} / \textbf{0.63} & 0.38 / 0.78 & 0.54 / 1.15 \\ 
&Ours-20               & 0.73 / 1.65 & \textbf{0.30} / \textbf{0.59} & 0.60 / 1.27 & 0.38 / 0.81 & \textbf{0.31} / \textbf{0.68} & \textbf{0.46} / \textbf{1.00} \\ \hline
\end{tabular}
\caption{Comparison of different methods on ETH (Column 3 and 4) and UCY datasets (Column 5-7). 
* We use a smaller test set on UNIV since 1 video is unable to download.}
\label{exp2}
\vspace{-5mm}
\end{table*}

\vspace{-2mm}
\subsection{Ablation Model}
In Table~\ref{multi-task}, we systematically evaluate our method through a series of ablation experiments, where ``ADE'' and ``FDE'' denotes the errors thus lower are better. ``Act'' is the mean Average Precision (mAP) of the activity label prediction over 29 activities and higher are better.

\noindent\textbf{Efficacy of rich visual features.} We investigate the feature contribution of person behavior and person interactions by separately ablating them. As shown in the first three rows in Table~\ref{multi-task}, both features are important to trajectory prediction while person behavior features are more essential for activity prediction. Individual feature ablations are shown in Table~\ref{exp-single}.

\noindent\textbf{Effect of focal attention.} In the fourth row of Table~\ref{multi-task}, we replace focal attention in Eq.~\eqref{eq:focal} with a simple average of the last hidden states from all encoders. Both trajectory and activity prediction hurt as a result.

\noindent\textbf{Impact of multi-task learning.} In the last three rows of Table~\ref{multi-task}, we remove the additional tasks of predicting the activity label or the activity location or both to see the impact of multi-task learning. Results show the benefit of our multi-task learning method.



\subsection{ETH \& UCY}\label{exp2-dataset}
\noindent\textbf{Dataset.} ETH~\cite{pellegrini2010improving} and UCY~\cite{lerner2007crowds} are common datasets for person trajectory prediction benchmark~\cite{alahi2016social,gupta2018social,manh2018scene,sadeghian2018sophie}.
Same as previous work~\cite{alahi2016social, gupta2018social,manh2018scene,sadeghian2018sophie}, we report performance by averaging over both datasets.
We use the same data processing method and settings detailed in~\cite{gupta2018social}. 
This benchmark includes videos from five scenes: ETH, HOTEL, UNIV, ZARA1 and ZARA2. Leave-one-scene-out data split is used and we evaluate our model on 5 sets of data. 
We follow the same testing scenario and baselines as in the previous section. We have also cited the latest state-of-the-art results from~\cite{sadeghian2018sophie}. Due to 1 video cannot be downloaded, we use a smaller test set for UNIV and a smaller training set across all splits. The other 4 test sub-datasets are the same as in~\cite{gupta2018social} so the numbers are comparable.

Since there is no activity annotation, we do not use activity label prediction module in our model. Since the annotation is only a point for each person and the human scale in each video doesn't change much, we apply a fixed size expansion from points for each video to get the person bounding box annotation for feature pooling. We do not use any other bounding box. We don't use any additional annotation compared to baselines to ensure a fair comparison.

\noindent\textbf{Implementation Details.} We do not use person keypoint feature. Final location loss and trajectory L2 loss are used. 
Unlike ~\cite{sadeghian2018sophie}, we don't utilize any data augmentation. We train our model for 40 epochs with the adadelta optimizer. Other hyper-parameters are the same as in Section~\ref{sec:exp-virat}.

\noindent\textbf{Results \& Analysis.} Experiments are shown in Table~\ref{exp2}. Our model outperforms other methods in both evaluations, where we obtain the best-published single model on ETH and best average performance on the ETH \& UCY benchmark. 
As shown in the table, our model performs much better on HOTEL and ZARA2. The average movement at each time-instant in these two scenes are 0.18 and 0.22, respectively, much lower than others: 0.389 (ZARA1), 0.460 (ETH), 0.258 (UNIV).
Recall that the leave-one-scene-out data split is used in training.
The results suggest other methods are more likely to overfit to the trajectories of large movements, e.g. Social-GAN~\cite{gupta2018social} often "over-shoot" when predicting the future trajectories. In comparison, our method uses attention to find the "right" visual signal and show better performance for trajectories of small movements on HOTEL and ZARA2 while still being competitive for trajectories of large movements.

\section{Conclusion}
In this paper, we have presented a new neural network model for predicting human trajectory and future activity simultaneously. We first encode a person through rich visual features capturing human behaviors and interactions with their surroundings. Then we add an auxiliary task of predicting the activity locations to facilitate the joint training process. We refer to the resulting model as \emph{\fancyname}. 
We showed the efficacy of our model on both popular and recent large-scale video benchmarks on person trajectory prediction.
In addition, we quantitatively and qualitatively demonstrated that our \emph{\fancyname} model successfully predicts meaningful future activities.

Our research goal is to promote human safety in applications such as robotics or autonomous driving. We experiment on the public benchmark ActEV, the primary driver of which is to support public safety and traffic monitoring and management by automatic activity detection in streaming video\footnote{{\footnotesize \url{https://actev.nist.gov/1B-Evaluation}}}. Our approach works on a predefined set of 30 activities provided by the NIST, such as ``loading'', ``object transfer''. See Table~\ref{tab:class} for the full list. Our system may not work beyond these predefined activities.



Future research into activity and path prediction may implicate ethical issues around privacy, safety and fairness and ought to be considered carefully before being used in real-world applications. Our method for predicting trajectory and activity has not been tested for different populations of people. As such, it is important to further evaluate these issues before employing the model in situations that may differentially impact people.


{
\bibliographystyle{ieee}
\bibliography{egbib}
}

\clearpage

\section*{Appendix}

\noindent In this appendix, we present more details and analysis for our experiments on the ActEV/VIRAT and ETH \& UCY Benchmarks. We also provide statistical comparisons of the two datasets.

\subsection{ActEV/VIRAT Details}

\subsubsection{Object \& Activity Class}
We show the object classes we used for our person interaction module and the activity classes for our activity prediction module in Table~\ref{tab:class}. Detailed class definition can be found on~{\footnotesize\url{https://actev.nist.gov/}}.

\subsubsection{Trajectory Type}
In ActEV/VIRAT dataset, there are two distinctive types of trajectory: relatively static and the moving ones. We label the person trajectory as moving if at time $T_{obs}$ there is an activity label of one of the following: "Walk", "Run", "Ride\_Bike", otherwise we label it as static trajectory.
Table~\ref{stats} shows the mean displacement in pixels between the last observed point and the prediction trajectory points. As we see, there is a large difference between the two types of trajectory. 

\subsubsection{Nearest Neighbor Experiment}
    Since the ActEV/VIRAT experiment is not camera-independent, we conduct a nearest neighbor experiment. Specifically, for each observed sequence in the test set, we use the nearest sequence in the training set as future predictions. As shown in Table~\ref{exp2}, it is non-trivial to predict human trajectory as people navigate differently even in the same scene.
    Please refer to the paper for evaluation metrics.

\subsubsection{Single Model Experiment}
    We train 20 identical \emph{\fancyname} models with different initialization for the single output experiment. We show the mean and standard deviation numbers in Table~\ref{exp2}.
    
\begin{table}[]
\centering
\begin{tabular}{c||c}
\hline
                     & Classes \\ \hline \hline
Object &   \makecell{ Bike, Construction\_Barrier, \\ Construction\_Vehicle,  Door, \\ Dumpster, Parking\_Meter, \\ Person, Prop, \\ Push\_Pulled\_Object, Vehicle }     \\ \hline
Activity  &  \makecell{ Carry, Close\_Door, Close\_Trunk, \\ Crouch, Enter, Exit, Gesture,\\ Interaction, Load,  Object\_Transfer,\\ Open\_Door,  Open\_Trunk, PickUp,\\ PickUp\_Person, Pull, Push, Ride\_Bike, \\ Run, SetDown, Sit, Stand, \\Talk, Talk\_phone, Texting, Touch,\\ Transport, Unload, Use\_tool, Walk}      \\ \hline
\end{tabular}
\caption{Object \& Activity Classes.}
\label{tab:class}
\end{table}

\begin{table}[]
\centering
\begin{tabular}{c||c|c}
\hline
                     & move\_traj & static\_traj \\ \hline \hline
Average Displacement (train) & 69.18      & 7.57         \\ 
Final Displacement (train)  & 124.79     & 14.63        \\
num\%  (train)              & 48.8\%    & 51.2\%      \\
Average Displacement (test)   & 75.78      & 12.01        \\
Final Displacement (test)  & 137.21     & 23.11        \\ 
num\%   (test)    & 61.9\%    & 38.1\%      \\ \hline
\end{tabular}
\label{stats}
\caption{Trajectory statistics for different trajectory class in ActEV dataset (on the training set).}
\end{table}

\subsubsection{Single Feature Ablation Experiments}
    We experiment with ablating  person-object, person-scene,  person keypoint and person  appearance feature, as shown in Table~\ref{exp-single}.
    
\subsection{Activity Detection Experiment} 
    Since we are predicting activities in the not so distant future, a system may perform well enough if it just outputs the current activity labels as the future prediction. We train an identical model to detect the activity labels at time $T_{obs}$ as the future prediction outputs, which results in a performance of  0.155 mAP for activity prediction and 18.27 ADE for trajectory prediction as shown in Table~\ref{exp-single}. Such a significant performance drop (0.192 vs. 0.155) suggests that activity prediction even for 4.8 seconds into the future is not a trivial task.

\subsubsection{More Qualitative Analysis}
\begin{figure*}[!t]
	\subfigure{\label{fig2:ourmodela}}
    \subfigure{\label{fig2:ourmodelb}}
    \subfigure{\label{fig2:ourmodelc}}
    \subfigure{\label{fig2:ourmodeld}}
    \subfigure{\label{fig2:ourmodele}}
    \subfigure{\label{fig2:ourmodelf}}
    \subfigure{\label{fig2:ourmodelg}}
    \subfigure{\label{fig2:ourmodelh}}
	\centering
		\includegraphics[width=1.0\textwidth]{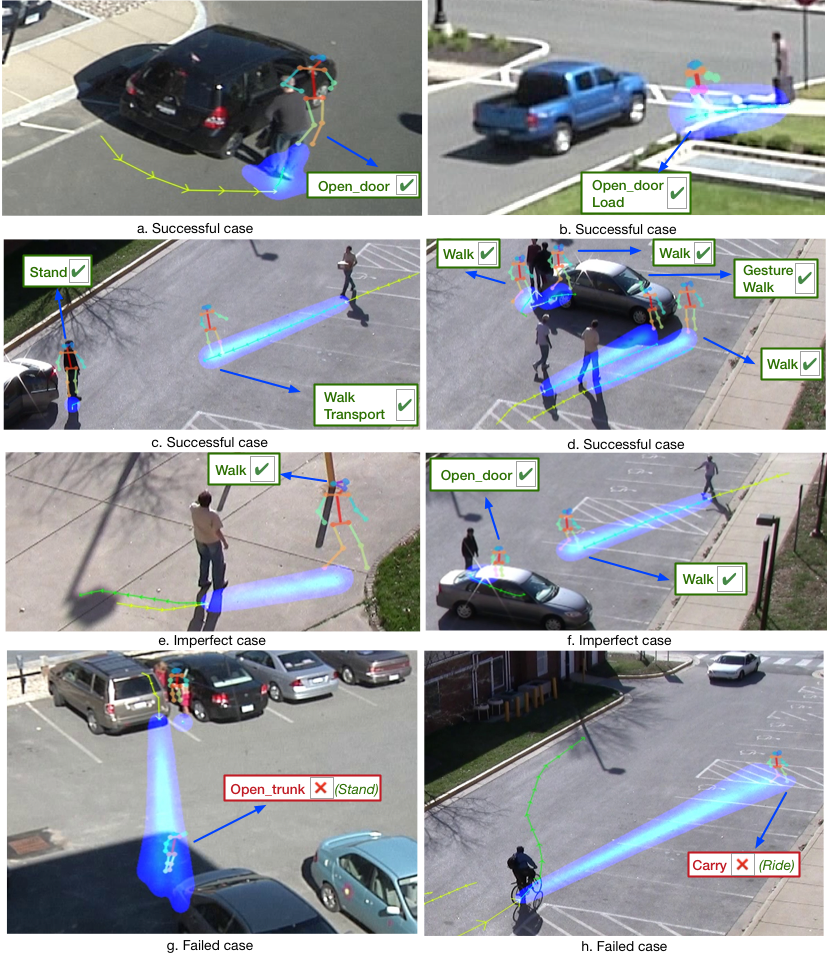}
	\caption{(Better viewed in color.) Qualitative analysis of our model.}  
	\label{fig:qualitative2}
\end{figure*}

We show more qualitative analysis in Fig.~\ref{fig:qualitative2}.
In each graph the yellow trajectories are the observable sequences of each person and the green trajectories are the ground truth future trajectories. The predicted trajectories are shown in the blue heatmap. To better visualize the predicted future activities of our method, we plot the person keypoint template for each predicted activity at the end of the predicted trajectory.

\noindent\textbf{Successful cases:} In Fig~\ref{fig2:ourmodela}, Fig~\ref{fig2:ourmodelb}, Fig~\ref{fig2:ourmodelc} and Fig~\ref{fig2:ourmodeld}, both the trajectory prediction and future activity prediction are correct. 
In Fig~\ref{fig2:ourmodeld}, our model successfully predicts the two persons at the bottom is going to walk past the car and also one of them is going to gesture at the other people by the trunk of the car.

\noindent\textbf{Imperfect cases:} In Fig~\ref{fig2:ourmodele} and Fig~\ref{fig2:ourmodelf}, although the activity predictions are correct, our model predicts the wrong trajectories. 
In Fig~\ref{fig2:ourmodele}, our model fails to predict that the person is going to the other direction.
In Fig~\ref{fig2:ourmodele}, our model fails to predict that the person near the car is going to open the front door instead of the back door.

\noindent\textbf{Failed cases:} In Fig~\ref{fig2:ourmodelg} and Fig~\ref{fig2:ourmodelh}, our model fails to predict both trajectories and activities. In Fig~\ref{fig2:ourmodelh}, the person on the bike is going to turn to avoid the incoming car while our model predicts a straight direction.

\subsubsection{Comparing ActEV/VIRAT to ETH \& UCY Benchmark} 
We compare the ActEV/VIRAT dataset and the ETH \& UCY trajectory benchmark in Table~\ref{comparison}. As we see, the ActEV/VIRAT dataset is much larger compared to the other benchmark. Also, the ActEV/VIRAT includes bounding box and activity annotations that could be used for multi-task learning. The ActEV/VIRAT is inherently different from the crow dataset since it includes diverse annotation of human activities rather than just passers-by, which makes trajectory prediction more purpose-oriented.
We show the trajectory numbers after processing based on the setting of eight-second-length sequences.
Note that in the public benchmark it is unbalanced since there is one crowded scene called "University" that contains over half of the trajectories in 4 scenes.

\begin{table}[t]
\centering
\begin{tabular}{l||c|c}
\hline
Metric & Nearest Neighbor & Our-Single-Model      \\ \hline \hline
ADE   &  40.04    & 17.99$\pm$0.043 \\
FDE   &  73.69      & 37.24$\pm$0.102 \\
{move\_ADE}  &	39.52  & 20.34$\pm$0.059 \\
{move\_FDE} & 72.67 & 42.54$\pm$0.146 \\ \hline
\end{tabular}
\caption{Our single model experiment on the ActEV/VIRAT benchmark.}
\label{exps2}
\end{table}

\begin{table}
\centering
\small
\begin{tabular}{l||c|c|c}
\hline
Method                             & ADE $\downarrow$  & FDE $\downarrow$  & Act mAP $\uparrow$\\ \hline
Our full model      & 17.91 & 37.11 & 0.192           \\ \hline
No p-object           & 18.17 & 37.13 & 0.198            \\ 
No p-scene          & 18.18 & 37.75    & 0.206            \\ 
No p-keypoint                & 18.25 & 37.96 & 0.190           \\ 
No p-appearance                        & 18.20 & 37.79 & 0.154                 \\ \hline
Act Detect                         & 18.27 & 37.68 & 0.155          \\ \hline
\end{tabular}
\caption{Single feature ablation \& activity detection experiments on the ActEV/VIRAT benchmark.}
\label{exp-single}
\end{table}

\begin{table}[]
\centering
\begin{tabular}{c||c|c}
\hline
               & ActEV       & ETH, UCY         \\ \hline 
\#Scene        & 5                   & 4                \\ \hline
Dataset Length & \makecell{4 hours \\22 minutes}  & 38 minutes       \\\hline
Resolutions    & \makecell{1920x1080,\\ 1280x720} & \makecell{640x480, \\720x576} \\ \hline
FPS            & 30                  & 25               \\\hline
\makecell{Annotation \\ FPS}            & 30                  & 2.5               \\\hline
\#Traj         & 84600               & \makecell{19359,\\  (10039 in Univ)}          \\ \hline
Annotations         & \makecell{Person+object \\ bounding boxes, \\ activities}             & \makecell{Person \\ coordinates}         \\ \hline
\end{tabular}
\label{comparison}
\caption{Comparison to commonly used person trajectory benchmark datasets.}
\end{table}

\subsection{ETH \& UCY Details}

\subsubsection{Dataset Difference Compared to SGAN}
The dataset we use is slightly different from the one in ~\cite{gupta2018social}, as some original videos are unavailable even though their trajectory annotations are provided. Specifically, two videos from UNIV scene, "students001", "uni\_examples", and one video from ZARA3, "crowds\_zara03", which is used in training for all corresponding splits in ~\cite{gupta2018social}, cannot be downloaded from the dataset website. Therefore, the test set for UNIV we use is smaller than previous methods~\cite{gupta2018social,sadeghian2018sophie} while the training set we use is about 34\% smaller.
Test sets for other 4 splits are the same therefore the numbers are comparable.

\subsubsection{Pre-Processing Details}
Since the annotation is only a point for each person and the human scale in each video doesn't change much, we apply a fixed size expansion from the annotated points for each video to get the person bounding box annotation for appearance and person-scene feature pooling.
Specifically, we use a bounding box size of 50 pixels by 80 pixels with the original annotation point putting at the center of the bottom line. 
All videos are resized to 720x576. The spatial dimension of the scene semantic segmentation feature is (64, 51) and two grid scales are used: (32, 26), (16, 13). 

\end{document}